\documentclass{article}

\usepackage{microtype}
\usepackage{graphicx}
\usepackage{subfigure}
\usepackage{booktabs} % for professional tables

\usepackage{amsfonts}
\usepackage{bm}
\usepackage{makecell}
\usepackage{dsfont}
\newcount\Comments % 0 suppresses notes to selves in text
\usepackage{array,multirow,graphicx}
\usepackage{color}
\usepackage{wrapfig}
\definecolor{darkgreen}{rgb}{0,0.5,0}
\definecolor{purple}{rgb}{1,0,1}
\newcommand{\kibitz}[2]{\ifnum\Comments=1\textcolor{#1}{#2}\fi}
\usepackage{stfloats}
\usepackage{hyperref}

\usepackage[accepted]{icml2019}

\icmltitlerunning{Active Multitask Learning with Committees}

\begin{document}

\twocolumn[
\icmltitle{Active Multitask Learning with Committees}




%%%% TODO author and affiliations
\icmlsetsymbol{equal}{*}

\begin{icmlauthorlist}
\icmlauthor{Jingxi Xu}{cu}
\icmlauthor{Da Tang}{cu}
\icmlauthor{Tony Jebara}{cu,netflix}
\end{icmlauthorlist}

\icmlaffiliation{cu}{Department of Computer Science, Columbia University, New
York, New York, USA}
\icmlaffiliation{netflix}{Netflix Inc., Los Gatos, California, USA}

\icmlcorrespondingauthor{Jingxi Xu}{jingxi.xu@columbia.edu}

\icmlkeywords{Machine Learning, ICML}

\vskip 0.3in
]



\printAffiliationsAndNotice{\icmlEqualContribution} % otherwise use the standard text.

\begin{abstract}
The cost of annotating training data has traditionally been a bottleneck for supervised learning approaches. The problem is further exacerbated when supervised learning is applied to a number of correlated tasks simultaneously since the amount of labels required scales with the number of tasks. To mitigate this concern, we propose an active multitask learning algorithm that achieves knowledge transfer between tasks. The approach forms a so-called \textit{committee} for each task that jointly makes decisions and directly shares data across similar tasks.  Our approach reduces the number of queries needed during training while maintaining high accuracy on test data. Empirical results on benchmark datasets show significant improvements on both accuracy and number of query requests.
\end{abstract}

% \begin{abstract}
% The expensive cost in annotating training data has been a bottleneck for many supervised learning models. Learning a number of correlated tasks simultaneously becomes a more pronounced problem as each task requires sufficient number of labelled data. We propose an active multitask learning algorithm that reduces the number of queries needed during training while maintaining a high accuracy on test data, by encouraging more knowledge transfer among tasks. This is achieved by having a \textit{committee} for each task to make joint decisions and directly sharing data among similar tasks. We empirically show the significant improvements on both accuracy and number of query requests on three benchmark multitask learning datasets.
% \end{abstract}

\section{Introduction}
A triumph of machine learning is the ability to predict with high accuracy. However, for the dominant paradigm, which is supervised learning, the main bottleneck is the need to annotate data, namely, to obtain labeled training examples. The problem becomes more pronounced in applications and systems which require a high level of personalization, such as music recommenders, spam filters, etc. Several thousand labeled emails are usually sufficient for training a good spam filter for a particular user. However, in real world email systems, the number of registered users is potentially in the millions, and it might not be feasible to learn a highly personalized spam filter for each of them by getting several thousand labeled data points for each user.

One method to relieve the need of the prohibitively large amount of labeled data is to leverage the relationship between the tasks, especially by transferring relevant knowledge from information-rich tasks to information-poor ones, which is called multitask learning in the literature. We consider multitask learning in an online setting where the learner sees the data sequentially, which is more practical in real world applications. In this setting, the learner receives an example at each time round, along with its task identifier, and then predicts its true label. Afterwards, the learner queries the true label and updates the model(s) accordingly.

The online multitask setting has received increasing attention in the machine learning community in recent years \cite{dekel2006online, abernethy2007multitask, dekel2007online, lugosi2009online, cavallanti2010linear, saha2011online, murugesan2016adaptive}. However, they make the assumption that the true label is readily available to be queried, which is impractical in many applications. Also, querying blindly can be inefficient when annotation is costly.

Active learning further reduces the work of the annotator by selectively requesting true labels from the oracles. Most approaches in active learning for sequential and stream-based problems adopt a measure of uncertainty\,/\,confidence of the learner in the current example \cite{cesa2006worst, cavallanti2009linear, orabona2011better, dekel2012selective,  agarwal2013selective}.  
%\citet{settles2012active} provides a more comprehensive survey on existing active learning scenarios and methods.

The recent work by \citet{murugesan2017active} combines active learning with online multitask learning using \textit{peers} or \textit{related tasks}. When the classifier of the current task is not confident, it first queries its similar tasks before requesting a true label from the oracle, incurring a lower cost. Their learner gives priority to the current task by always checking its confidence first. In the case when the current task is confident, the opinions of its peers are ignored. 

This paper proposes an active multitask learning framework which is more \textit{humble}, in a sense that both the current task and its peers' predictions are considered simultaneously using a weighted sum. We have a \textit{committee} which makes joint decisions for each task. In addition, after the true label of a training sample is obtained, this sample is shared directly to similar tasks, which makes training more efficient.

% Our approach maintains a \textit{committee} for each task, where the weights of each expert (classifier) are automatically learned and adapted. In the end, the weights also reflect the similarities between different tasks. When the true label is queried, this example is directly shared to similar tasks, which makes training more efficient.

\section{Problem Formulation}
The problem formulation and setup are similar to \cite{murugesan2016adaptive, murugesan2017active}. Suppose we are given $K$ tasks and the $k$-th task is associated with $N_k$ training samples. We consider each task to be a linear binary classification problem, but the extensions to multi-class or non-linear cases are straightforward. We use the good-old perceptron-based update rule in which the model for a given task is only updated when the prediction for that training example is in error. The data for task $k$ is $\{x^{(i)}_k, y^{(i)}_k\}^{N_k}_{i=1}$, where $x^{(i)}_k \in \mathbb{R}^{D}$ is the $i$-th instance from the $k$-th task, $y^{(i)}_k \in \{-1, +1\}$ is the corresponding label and $D$ is the dimension of features. When the notation is clear from the context, we drop task index $k$ and simply write $((x^{(i)},\, k), \, y^{(i)})$. We consider the online setting where the training example $((x^{(t)},\, k), \, y^{(t)})$ comes at round $t$. 

Denote $\{w_k^{(t)}\}_{k \in [K]}$ the set of weights learned for the $K$ binary classifiers at round $t$. Also denote $\bm{w} \in \mathbb{R}^{K\times D}$ the weight matrix whose $k$-th row is $w_k$. The label $\hat{y}^{(t)}$ is predicted based on the sign of the output value from the model. Then the hinge loss of task $k$ on the sample $((x^{(t)},\, k), \, y^{(t)})$ at round $t$ is given by $\ell_{kk}^{(t)} = \left( 1 - y^{(t)} \langle x^{(t)}, w_k^{(t)} \rangle \right)_+$. In addition, we also consider the losses of its peer tasks $m \, (m \neq k)$ as $\ell_{km}^{(t)} = \left( 1 - y^{(t)} \langle x^{(t)}, w_m^{(t)} \rangle \right) _+$. $\ell_{km}^{(t)}$ indicates the loss incurred by using task $m$'s knowledge\,/\,classifier to predict the label of task $k$'s training sample. $\ell_{km}^{(t)}$ plays an important role in learning the similarities among tasks and hence the committee weights. Intuitively, two tasks should be more similar if one task's training samples can be correctly predicted using the other task's classifier.

The goal of this paper is to achieve a high accuracy on the test data, and at the same time to issue as small a number of queries to the oracle as possible during training, by efficiently sharing and transferring knowledge among similar tasks. 
%\jx{information can be shared!}

\section{Active Multitask Learning w/ Committees}

\begin{algorithm}[t]
    \small
    \caption{Active Multitask Learning with Committees} 
    \label{alg:amlc}
    \begin{algorithmic}[1]
        \FUNCTION{AMLC \,($b$, $C$, $T$)}
        \STATE Initialize $w_m^{(0)} = \bm0_D, \forall m \in [K], \bm{\tau}^{(0)} = \frac{1}{K}\bm1_{K\times K}$
        \FOR{$t = 1,2,..,T$}
            \STATE Receive $(x^{(t)}, k)$
            \STATE Compute $p_{km}^{(t)} = \langle x^{(t)}, w_m^{(t-1)} \rangle$ for $m \in [K]$
            \STATE $p = \sum_{m \in [K]} p_{km}^{(t)} \tau_{km}^{(t-1)}$
            \STATE Predict $\hat{y}^{(t)} = \textrm{sign}(p)$
            \STATE Draw  $P^{(t)}\sim\textrm{Bernoulli}\left(\frac{b}{b+|p|}\right)$.
            \IF{$P^{(t)} = 1$}
                \STATE Query true label $y^{(t)}$ and set $M^{(t)} = \mathds{1}\left[ y^{(t)} \neq \hat{y}^{(t)} \right]$
                \STATE Update $w_k^{(t)} = w_k^{(t-1)}+ P^{(t)} M^{(t)}           y^{(t)}x^{(t)}$ 
                \STATE Update $\bm{\tau}$: 
                    $$\tau_{km}^{(t)} = \frac{\tau_{km}^{(t-1)} e^{-C \cdot \frac{l_{km}^{(t)}}{\lambda} }}{\sum_{m'\in[K]} \tau_{km'}^{(t-1)} e^{-C \cdot \frac{l_{km'}^{(t)}}{\lambda} }}, m\in [K] $$
                \FOR{$\forall m \in [K]$, and $m \neq k$}
                    \STATE Set $S_m^{(t)} = \mathds{1}\left[ \textrm{sign}\left(p_{km}^{(t)}\right) \neq \hat{y}^{(t)} \wedge \tau_{km}^{(t)} \geq \tau_{kk}^{(t)} \right]$
                    \STATE Update $w_m^{(t)} = w_m^{(t-1)}+ S_m^{(t)} y^{(t)}x^{(t)}$
                \ENDFOR
            \ENDIF
        \ENDFOR
        \STATE \textbf{return} $\bm{\tau}^{(t)} \bm{w}^{(t)}$
        \ENDFUNCTION
    \end{algorithmic}
\end{algorithm}

In this section we introduce our algorithm \textit{Active Multitask Learning with Committees} (AMLC) as shown in Algorithm~\ref{alg:amlc}. This algorithm provides an efficient way for online multitask learning. Each task uses not only its own knowledge but also knowledge from other tasks, and shares training examples across similar tasks when necessary. The two main components of Algorithm~\ref{alg:amlc} are described in Section~\ref{sec:committee} and \ref{sec:share}. In Section~\ref{sec:compare}, we compare AMLC with the state-of-the-art online multitask learning algorithm.

\subsection{Learning with Joint Decisions}
\label{sec:committee}
We maintain and update a \textit{relationship matrix} $\bm{\tau} \in \mathbb{R}^{K\times K}$ through the learning process. The $k$-th row of $\bm{\tau}$, denoted $\tau_k$, is the \textit{committee weight vector} for task $k$, also referred to as \textit{committee} for brevity. Element $\tau_{ij}$ of the relationship matrix indicates the closeness or similarity between task $i$ and task $j$, and also the importance of task $j$ in task $i$'s committee in predicting. Given a sample $((x^{(t)},\, k), \, y^{(t)})$ at round $t$, the confidence of task $k$ is jointly decided by its committee; namely, a weighted sum of confidences of all tasks, $p = \sum_{m \in [K]} p_{km}^{(t)} \tau_{km}^{(t-1)}$, where $p_{km}^{(t)} = \langle x^{(t)}, w_m^{(t-1)} \rangle$ for $m \in [K]$. Each confidence is just the common confidence measure for perceptron, using distance from the decision boundary \cite{cesa2006worst}. The prediction is done by taking the sign of the confidence value. The learner then makes use of this confidence value by drawing a sample $P^{(t)}$ from a Bernoulli distribution, to decide whether to query the true label of this sample. The larger $p$ is, the more likely for $P^{(t)}$ to be 0, signifying greater confidence. The hyperparameter $b$ controls the level of confidence that the current task has to have to not request the true label.

The learner only queries the true label when the current task's committee turns out to be unconfident. Another binary variable $M^{(t)}$ is set to be 1 if task $k$ makes a mistake. Subsequently, its weight vector is updated following the conventional perceptron scheme. The learner then updates the relationship matrix following a similar policy as in \cite{murugesan2016adaptive, murugesan2017active}. For tasks that incur no loss on this example, their weights in the committee are not changed. On the other hand, for tasks that incur non-zero loss, their weights are decreased by a factor $\exp ({-C \cdot l_{km}^{(t)}} / \lambda)$. The hyperparameter $C$ decides how much decrease happens on the weight given non-zero loss, and $\lambda = \sum_{m=1}^K l_{km}^{(t)}$. These new weights are then normalized to sum to 1. 

\subsection{Data Sharing}
\label{sec:share}

\begin{table*}[tp]
\centering
\resizebox{\textwidth}{!}{
\begin{tabular}{ccccccc}
    \toprule
    \textbf{Models} & \multicolumn{2}{c}{\textbf{Landmine Detection}} & \multicolumn{2}{c}{\textbf{Spam Detection}} & \multicolumn{2}{c}{\textbf{Sentiment}} \\
    
    & \textit{Accuracy} & \textit{\#Queries} & \textit{Accuracy} & \textit{\#Queries} & \textit{Accuracy} & \textit{\#Queries} \\
    \midrule
    
    Random & \makecell{$0.8914 \,\scriptstyle \pm 0.0126$} & \makecell{$2323.5 \scriptstyle \,\pm 11.5$} & \makecell{$0.7940 \scriptstyle \,\pm 0.0131$} & \makecell{$751.8 \scriptstyle \,\pm 14.3$} & \makecell{$0.6068 \scriptstyle  \,\pm 0.0065$} & \makecell{$1092.0 \scriptstyle \,\pm 16.4$}\\

    Independent & \makecell{$0.9070 \,\scriptstyle \pm 0.0079$} & \makecell{$2770.3 \,\scriptstyle \pm 25.8$} & \makecell{$0.8232 \,\scriptstyle \pm 0.0159$} & \makecell{$1188.6 \,\scriptstyle \pm 6.6$} & \makecell{$0.6404 \,\scriptstyle \pm 0.0050$} & \makecell{$1987.5 \,\scriptstyle \pm 7.1$}\\

    PEER & \makecell{$0.9362 \,\scriptstyle \pm 0.0025$} & \makecell{$1206.0\,\scriptstyle \pm 23.2$}& \makecell{$0.8334\,\scriptstyle \pm 0.0134$} & \makecell{$1085.7\,\scriptstyle \pm 13.9$} & \makecell{$0.6425\,\scriptstyle \pm 0.0067$} & \makecell{$1979.7\,\scriptstyle \pm 10.3$}\\

    PEER+Share & \makecell{$0.9231\,\scriptstyle\pm 0.0112$} & \makecell{$1885.8\,\scriptstyle\pm 71.3$} & \makecell{$\bm{0.8766 \,\scriptstyle\pm 0.0135}$} & \makecell{$935.3\,\scriptstyle\pm 18.3$} & \makecell{$\bm{0.7645\,\scriptstyle\pm 0.0077}$} & \makecell{$1754.5\,\scriptstyle\pm 9.2$}\\
    
    AMLC & \makecell{$\bm{0.9367\,\scriptstyle\pm 0.0020}$} & \makecell{$\bm{189.1\,\scriptstyle\pm 12.0}$} & \makecell{$\bm{0.8706\,\scriptstyle\pm 0.0102}$} & \makecell{$\bm{321.3\,\scriptstyle\pm 15.6}$} & \makecell{$0.7358\,\scriptstyle\pm 0.0093$} & \makecell{$\bm{633.9\,\scriptstyle\pm 11.4}$}\\
    \bottomrule
    
\end{tabular}}
\caption{Accuracy on test set and total number of queries during training over 10 random shuffles of the training examples. The $95\%$ confidence level is provided after the average accuracy. The best performance is highlighted in bold. On Spam Detection, both PEER+Share and AMLC are highlighted because AMLC has a lower mean but also smaller variance.}
\label{tab:results}
\end{table*}

To further encourage data sharing and information transfer between similar tasks, after the true label is obtained, the learner also shares the data with similar tasks of task $k$, so that peer tasks can learn from this sample as well. Similar tasks are identified by having a larger weight than the current task in the committee. We set $S^{(t)}_m = 1$ to indicate task $m$ is a similar task to $k$ and thus the data is shared with it.

\subsection{Comparison with PEER}
\label{sec:compare}
The most related work to ours is \textit{active learning from peers} (PEER) \cite{murugesan2017active}. In this section we discuss the main difference between our method and theirs with some intuition. 

Firstly, we do not treat the task itself and its peer tasks separately. Instead, the final confidence of the current task is jointly decided using the confidences of all tasks, weighted by the committee weight vector. It is humble in a sense that it always considers its peer tasks' advice when making a decision. There are two main advantages of our approach. 1) For PEER, no updates happen and no knowledge is transferred when the current task itself is confident. This can result in difficulties for the learner to recover from being blindly confident. Blind confidence happens when the classifier makes mistakes on training examples but with high confidence, especially in early stage of training when data are not enough. %At the end, the algorithm outputs $\bm{\tau}^{(t)} \bm{w}^{(t)}$, \jx{unclear?} so that in the testing period, the prediction is still made jointly by committee. 
2) Our method updates the committee weight vector while keeping $\sum_{m\in[K]} \tau_{km} = 1$ instead of $\sum_{m\in[K], m\neq k} \tau_{km} = 1$. It then becomes possible that the current task itself has an equal or lower influence than other tasks on the final prediction. This is more desirable because identical tasks should have equal weights, and information-poor tasks should rely more on their information-rich peers when making predictions.

Secondly, our algorithm enables the sharing of training data across similar tasks directly, after acquiring the true label of this data. Querying can be costly, and the best way to make use of the expensive label information is to share it. Assuming that all tasks are identical, the most productive algorithm would merge all data to learn a single classifier. PEER is not able to achieve this because each task is still trained independently, since tasks only have access to their own data. Though PEER indirectly accesses others' data through querying peer tasks, this sharing mechanism can be insufficient when tasks are highly similar. In the case that all tasks are identical, our algorithm converges to a relationship matrix with identical elements and eventually all tasks are trained on every example that has been queried.

\section{Experiments}

\begin{figure*}[t]
\begin{center}
    \includegraphics[width=\textwidth]{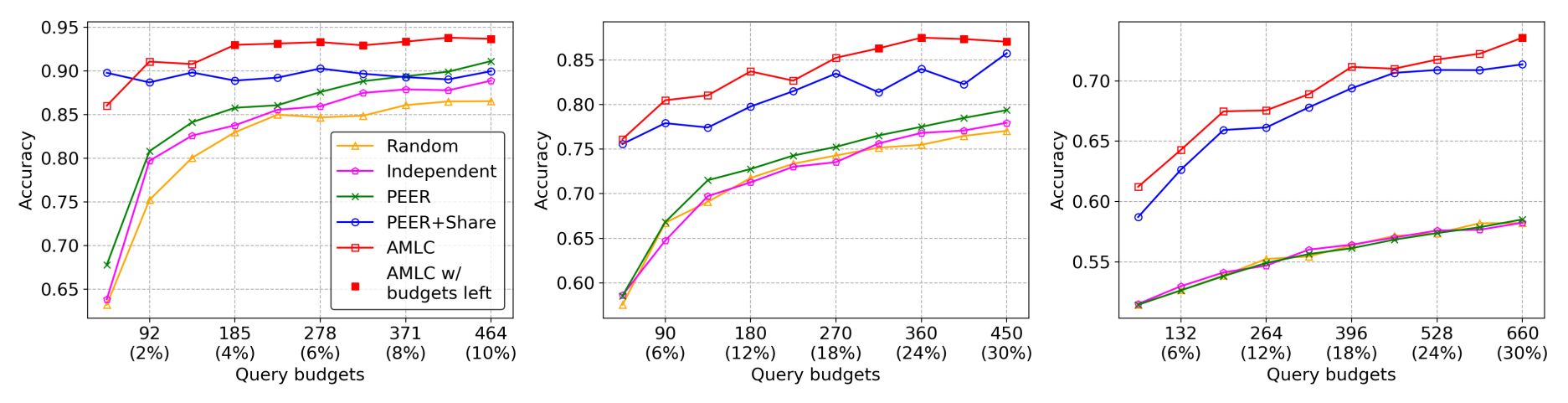}
\end{center}
\vskip -0.1in
\caption{Accuracy on test set w.r.t. query budget. Left, middle and right are Landmine Detection, Spam Detection and Sentiment respectively. The filled markers indicate that there are still queries left in the budget.}
\label{fig:plots}
\end{figure*}

In this section, we evaluate our proposed algorithm on three benchmark datasets for multitask learning, and compare our performance with many baseline models.
% \footnote{Code is available at \url{xxxxxxxxxxxxxxxxxxxx}}. 
We set $b = 1$ for all the experiments and tune the value of $C$ from 20 values using 10-fold cross validation. Unless otherwise specified, all other model parameters are chosen via 10-fold cross validation.

\subsection{Benchmark Datasets}
\textit{Landmine Detection}\footnote{\url{http://www.ee.duke.edu/~lcarin/LandmineData.zip}} consists of 19 tasks collected from different landmine fields. Each task is a binary classification problem: landmines (+) or clutter (-), and each example consists of 9 features. 
\textit{Spam Detection}\footnote{\url{http://ecmlpkdd2006.org/challenge.html}} consists of labeled training data: spam (+) or non-spam (-) from the inboxes of 15 users, and each user is considered as a single task. 
\textit{Sentiment Analysis}\footnote{\url{http://www.cs.jhu.edu/~mdredze/datasets/sentiment/}} \cite{blitzer7domain} consists of product reviews from Amazon containing reviews from 22 domains. We consider each domain as a binary classification task: positive review (+) and negative review (-). Details about our training and test sets are shown in Appendix~\ref{appendix:dateset}.

\subsection{Results}
We compare the performance of $5$ different models. \textit{Random} does not use any measure of confidence. Namely, the probability of querying or not querying true label are equal. \textit{Independent} uses the confidence which is purely computed form the weight vector of the current task. Obviously both Random and Independent have no knowledge transfer among tasks. \textit{PEER} is the algorithm from \cite{murugesan2017active}. \textit{AMLC} (Active Multitask Learning with Committees) is our proposed method as shown in Algorithm~\ref{alg:amlc}. In addition, we also show the performance of \textit{PEER+Share}, in which we simply add to PEER the data sharing mechanism as illustrated in section~\ref{sec:share}.

Table~\ref{tab:results} shows the accuracy on test set and the total number of queries (label requests) to oracles during training of five models. Each value is the average of 10 random shuffles of the training set. The $95\%$ confidence level is also shown. Notice that our re-implementation of PEER achieves similar performance on the Landmine and Spam datasets but seems to perform worse on Sentiment. The reason is that we are using a different representation of the training examples. We use the default bag-of-words representation coming with the dataset and there are approximately $2.9M$ features.

The highlighted values illustrate the best performance across all models. On Spam Detection, AMLC is also highlighted because it is more confident about its accuracy even though the actual value is slightly lower than PEER+Share. It can be seen that our proposed methods (PEER+Share and AMLC) significantly outperform the the others. PEER has better performance compared to Random and Independent but still behaves worse than PEER+Share and AMLC. It can be shown that simply adding data sharing can improve both accuracy and number of queries used during training. The only exception is on Landmine Detection, where PEER+Share requests more queries than PEER.  Though simply adding data sharing results in improvement, after learning with joint decisions in AMLC, we observe further drastic decrease on the number of queries, while maintaining a high accuracy.

Another goal of active multitask learning is to efficiently make use of the labels. In order to evaluate this, we give each model a fixed number of query budget and the training process is ended after the budget is exhausted. We show three plots (one for each dataset) in Figure~\ref{fig:plots}. Based on the difficulty of learning from each dataset, we choose different budgets to evaluate (up to $10\%, 30\%$ and $30\%$ of the total training examples for Landmine, Spam and Sentiment respectively). We can see that given a limited number of query budgets, AMLC outperforms all models on all three datasets, as a result of it encouraging more knowledge transfer among tasks. It is worth noting that the Landmine dataset is quite unbalanced (high proportion of negative labels), and PEER+Share and AMLC can achieve high accuracy with extremely limited number of queries. However, the classifier learned by PEER+Share is unconfident and thus it keeps requesting true labels in the following training process.

\section{Conclusion}
We propose a new active multitask learning algorithm that encourages more knowledge transfer among tasks compared to the state-of-the-art models, by using joint decision\,/\,prediction and directly sharing training examples with true labels among similar tasks. Our proposed methods achieve both higher accuracy and lower number of queries on three benchmark datasets for multitask learning problems. Future work includes theoretical analysis of the error bound and comparison with those of the baseline models. Another interesting direction is to handle unbalanced task data. In other words, one task has much more\,/\,less training data than the others.


\bibliography{example_paper}

\begin{thebibliography}{14}
\providecommand{\natexlab}[1]{#1}
\providecommand{\url}[1]{\texttt{#1}}
\expandafter\ifx\csname urlstyle\endcsname\relax
  \providecommand{\doi}[1]{doi: #1}\else
  \providecommand{\doi}{doi: \begingroup \urlstyle{rm}\Url}\fi

\bibitem[Abernethy et~al.(2007)Abernethy, Bartlett, and
  Rakhlin]{abernethy2007multitask}
Abernethy, J., Bartlett, P., and Rakhlin, A.
\newblock Multitask learning with expert advice.
\newblock In \emph{International Conference on Computational Learning Theory},
  pp.\  484--498. Springer, 2007.

\bibitem[Agarwal(2013)]{agarwal2013selective}
Agarwal, A.
\newblock Selective sampling algorithms for cost-sensitive multiclass
  prediction.
\newblock In \emph{International Conference on Machine Learning}, pp.\
  1220--1228, 2013.

\bibitem[Blitzer et~al.()Blitzer, Dredze, and Pereira]{blitzer7domain}
Blitzer, J., Dredze, M., and Pereira, F.
\newblock Domain adaptation for sentiment classification.
\newblock In \emph{45th Annv. Meeting of the Assoc. Computational Linguistics
  (ACL’07)}.

\bibitem[Cavallanti et~al.(2009)Cavallanti, Cesa-Bianchi, and
  Gentile]{cavallanti2009linear}
Cavallanti, G., Cesa-Bianchi, N., and Gentile, C.
\newblock Linear classification and selective sampling under low noise
  conditions.
\newblock In \emph{Advances in Neural Information Processing Systems}, pp.\
  249--256, 2009.

\bibitem[Cavallanti et~al.(2010)Cavallanti, Cesa-Bianchi, and
  Gentile]{cavallanti2010linear}
Cavallanti, G., Cesa-Bianchi, N., and Gentile, C.
\newblock Linear algorithms for online multitask classification.
\newblock \emph{Journal of Machine Learning Research}, 11\penalty0
  (Oct):\penalty0 2901--2934, 2010.

\bibitem[Cesa-Bianchi et~al.(2006)Cesa-Bianchi, Gentile, and
  Zaniboni]{cesa2006worst}
Cesa-Bianchi, N., Gentile, C., and Zaniboni, L.
\newblock Worst-case analysis of selective sampling for linear classification.
\newblock \emph{Journal of Machine Learning Research}, 7\penalty0
  (Jul):\penalty0 1205--1230, 2006.

\bibitem[Dekel et~al.(2006)Dekel, Long, and Singer]{dekel2006online}
Dekel, O., Long, P.~M., and Singer, Y.
\newblock Online multitask learning.
\newblock In \emph{International Conference on Computational Learning Theory},
  pp.\  453--467. Springer, 2006.

\bibitem[Dekel et~al.(2007)Dekel, Long, and Singer]{dekel2007online}
Dekel, O., Long, P.~M., and Singer, Y.
\newblock Online learning of multiple tasks with a shared loss.
\newblock \emph{Journal of Machine Learning Research}, 8\penalty0
  (Oct):\penalty0 2233--2264, 2007.

\bibitem[Dekel et~al.(2012)Dekel, Gentile, and Sridharan]{dekel2012selective}
Dekel, O., Gentile, C., and Sridharan, K.
\newblock Selective sampling and active learning from single and multiple
  teachers.
\newblock \emph{Journal of Machine Learning Research}, 13\penalty0
  (Sep):\penalty0 2655--2697, 2012.

\bibitem[Lugosi et~al.(2009)Lugosi, Papaspiliopoulos, and
  Stoltz]{lugosi2009online}
Lugosi, G., Papaspiliopoulos, O., and Stoltz, G.
\newblock Online multi-task learning with hard constraints.
\newblock \emph{arXiv preprint arXiv:0902.3526}, 2009.

\bibitem[Murugesan \& Carbonell(2017)Murugesan and
  Carbonell]{murugesan2017active}
Murugesan, K. and Carbonell, J.
\newblock Active learning from peers.
\newblock In \emph{Advances in Neural Information Processing Systems}, pp.\
  7011--7020, 2017.

\bibitem[Murugesan et~al.(2016)Murugesan, Liu, Carbonell, and
  Yang]{murugesan2016adaptive}
Murugesan, K., Liu, H., Carbonell, J., and Yang, Y.
\newblock Adaptive smoothed online multi-task learning.
\newblock In \emph{Advances in Neural Information Processing Systems}, pp.\
  4296--4304, 2016.

\bibitem[Orabona \& Cesa-Bianchi(2011)Orabona and
  Cesa-Bianchi]{orabona2011better}
Orabona, F. and Cesa-Bianchi, N.
\newblock Better algorithms for selective sampling.
\newblock In \emph{International conference on machine learning}, pp.\
  433--440. Omnipress, 2011.

\bibitem[Saha et~al.(2011)Saha, Rai, Daum{\~A}, Venkatasubramanian,
  et~al.]{saha2011online}
Saha, A., Rai, P., Daum{\~A}, H., Venkatasubramanian, S., et~al.
\newblock Online learning of multiple tasks and their relationships.
\newblock In \emph{Proceedings of the Fourteenth International Conference on
  Artificial Intelligence and Statistics}, pp.\  643--651, 2011.

\end{thebibliography}
\bibliographystyle{icml2019}

\newpage
\appendix
{\Large \textbf{Appendix}}

We describe in details about the three datasets and the train\,/\,test split used for the experiments. The description is adapted from \cite{murugesan2016adaptive, murugesan2017active}.

\section{Dataset Details}
\label{appendix:dateset}

\paragraph{Landmine Detection} consists of 19 tasks collected from different landmine fields. Each task is a binary classification problem: landmines (+) or clutter (-) and each example consists of 9 features extracted from radar images with four moment-based features, three correlation-based features, one energy ratio feature and a spatial variance feature.  Landmine data is collected from two different terrains: tasks 1-10 are from highly foliated regions and tasks 11-19 are from desert regions; therefore tasks naturally form two clusters. Any hypothesis learned from a task should be able to utilize the information available from other tasks belonging to the same cluster.

\paragraph{Spam Detection} is obtained from ECML PAKDD 2006 Discovery challenge for the spam detection task.  We used the task B challenge dataset which consists of labeled training data from the inboxes of 15 users.  We consider each user as a single task and the goal is to build a personalized spam filter for each user.  Each task is a binary classification problem:  spam (+) or non-spam (-) and each example consists of approximately $150K$ features representing term frequency of the word occurrences. Since some spam is universal to all users (e.g. financial scams), some messages might be useful to certain affinity groups, but spam to most others. Such adaptive behavior of user’s interests and dis-interests can be modeled efficiently by utilizing the data from other users to learn per-user model parameters.

\paragraph{Sentiment Analysis} contains product reviews from many domains on Amazon \cite{blitzer7domain}. We consider each domain as a binary classification task. Reviews with rating $>$ 3 are labeled positive (+), those with rating $<$ 3 are labeled negative (-), and reviews with rating $=$ 3 are discarded as the sentiments are ambiguous and hard to predict.  We choose 22 domains which have enough data for both positive and negative labels. We use the default preprocessed bag-of-words representation that comes with the dataset and each example consists of approximately $2.9M$ features.

We choose 3040 examples (160 training examples per task) for landmine, 1500 emails for spam (100 emails per user inbox) and 2200 reviews for sentiment (100 reviews per domain) for our experiments. For landmine and spam, we use the rest of the examples for test set, and for sentiment, we select another 300 (using all remaining data if not enough) examples for test set. On average, each task in landmine, spam, sentiment has 509, 400 and 395 examples respectively. Note that we intentionally keep the size of the training data small to drive the need for learning from other tasks, which diminishes as the training set per task becomes larger.
%%%%%%%%%%%%%%%%%%%%%%%%%%%%%%%%%%%%%%%%%%%%%%%%%%%%%%%%%%%%%%%%%%%%%%%%%%%%%%%
%%%%%%%%%%%%%%%%%%%%%%%%%%%%%%%%%%%%%%%%%%%%%%%%%%%%%%%%%%%%%%%%%%%%%%%%%%%%%%%

\end{document}